\documentclass{article}
\usepackage{spconf,amsmath,graphicx}
\usepackage{color}
\usepackage{siunitx}

\graphicspath{{Figures/}}

\title{RESTORATION OF MARKER OCCLUDED HEMATOXYLIN AND EOSIN STAINED WHOLE SLIDE HISTOLOGY IMAGES USING GENERATIVE ADVERSARIAL NETWORKS}

\name{Bairavi Venkatesh$^{\star}$, Tosha Shah$^{\dagger}$, Antong Chen$^{\star}$, Soheil Ghafurian$^{\dagger,\pounds}$}

\address{
$^{\star}$Image Data Analytics, Informatics IT, Merck \& Co., Inc., West Point, PA 19486, USA\\
$^{\dagger}$Image Data Analytics, Informatics IT, Merck \& Co., Inc., Rahway, NJ 07065, USA\\
$^{\pounds}$soheil.ghafurian@merck.com\\
}
\begin{document}
%
\maketitle
\begin{abstract}

It is common for pathologists to annotate specific regions of the tissue, such as tumor, directly on the glass slide with markers.
Although this practice was helpful prior to the advent of histology whole slide digitization, it often occludes important details which are increasingly relevant to immuno-oncology due to recent advancements in digital pathology imaging techniques. 
The current work uses a generative adversarial network with cycle loss to remove these annotations while still maintaining the underlying structure of the tissue by solving an image-to-image translation problem. 
We train our network on up to 300 whole slide images with marker inks and show that 70\% of the corrected image patches are indistinguishable from originally uncontaminated image tissue to a human expert. 
This portion increases 97\% when we replace the human expert with a deep residual network. 
We demonstrated the fidelity of the method to the original image by calculating the correlation between image gradient magnitudes. We observed a revival of up to 94,000 nuclei per slide in our dataset, the majority of which were located on tissue border.

\end{abstract}
\section{Introduction}
\label{sec:intro}

\begin{figure*}[htb]
  \centering
  \centerline{\includegraphics[width=\textwidth]{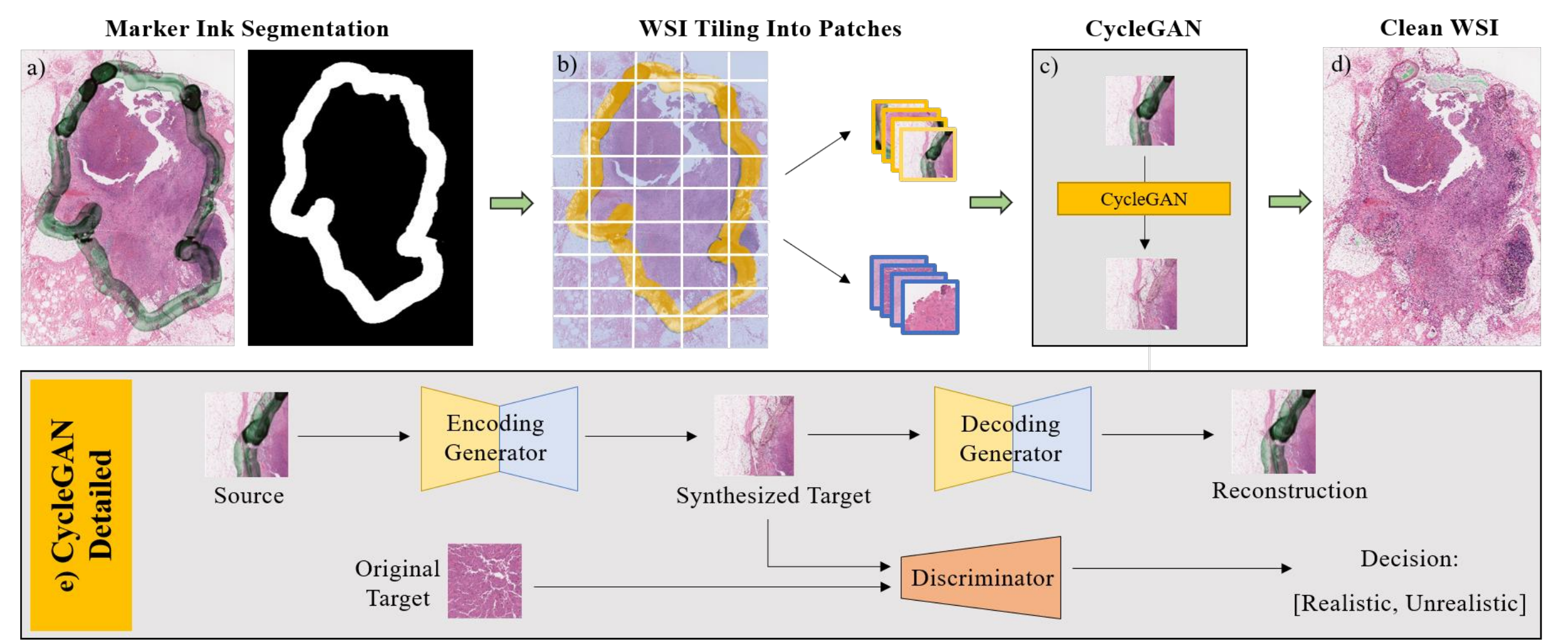}}
\caption{
The pipeline employed for training the model
}
\label{fig:framework}
\end{figure*}

It is a common practice for pathologists to annotate regions of interest, such as tumor, directly on the glass slide to which the tissue sample was affixed. 
These annotations are useful for tasks at the macro level such as tumor detection and routine cancer diagnosis. 
Though the rise of powerful commercial scanners coupled with advanced image viewers has enabled widespread slide digitization, many legacy slides are still scanned with the pathologist's original hand-drawn annotations. 
Such annotations often occlude important details of the image, preventing implementation of recent digital pathology workflows at the micro level. 
For example, on-going research has shown the significance of the tumor micro-environment on cancer progression and treatment, specifically in the field of immuno-oncology~\cite{io1, io2}. 
The identification, localization, and spatial relationships of certain immune cells with tumor cells hold valuable prognostic potential for personalized treatment regimens~\cite{io3}. 
However, tasks such as nuclei segmentation and classification, and the extraction of features and predictive biomarkers in the following steps cannot be accurately conducted without first producing an occlusion-free image.


CycleGAN was introduced in 2017~\cite{CycleGAN} for unpaired image translation tasks, in which the training images in the source and target domains do not need to correspond to one another. 
This method uses cycle consistency to avoid mode collapse, which is the transformation of all input images in the source domain to one or a few images in the target domain.
Through adding a second generator convolutional neural network (CNN), CycleGAN ensures that the individual information of each image is maintained throughout the transformation by reconstructing the original image from the output.

CycleGAN is typically good at translating the texture of the input to match the target while not altering the structure significantly.
This has motivated many applications for CycleGAN including digital pathology, where it has been used for data augmentation and style transfer.
Fu et al.~\cite{fu2018fluorescence} used this model to transform artificial nuclei segmentation masks to synthetic nuclei fluorescence images, thus producing a large annotated dataset of nuclei required for training a nuclei segmentation CNN.
Mahmood et al.~\cite{mahmood2018deep} developed a similar approach for generating training datasets for nuclei segmentation in hematoxylin and eosin stained (H\&E) images.
Additionally, they trained a CycleGAN to imitate nuclei segmentation that was performed by human experts.
Shaban et al.~\cite{staingan} showed that CycleGAN outperforms state-of-the-art methods for stain normalization, in which the goal is to eliminate the image variations resulting from different imaging parameters such as equipment and environment.
Xu et al.~\cite{xu2019gan} took it a step further and used CycleGAN for transforming H\&E images into immunohistochemistry images in order to take advantage of possibilities that these images provide.

In this paper, we restore the marker-occluded regions of H\&E images to clean ones.
We treat this as a style transfer problem and train a CycleGAN with up to 300 whole slide images (WSI) to remove the marker ink (Figure~\ref{fig:framework}).
We show the quality of the results by performing a blind test once by a human expert and once by a separately trained deep residual network.
We also demonstrate the fidelity of our results by conducting a morphological test of the results.
Finally, we process a sample of the reconstructed WSIs and attain an increase of up to 94,000 detected nuclei per slide.

\section{Method}
\subsection{Data preparation and training}
\begin{figure}[htb]
  \centering
  \centerline{\includegraphics[width=8.5cm]{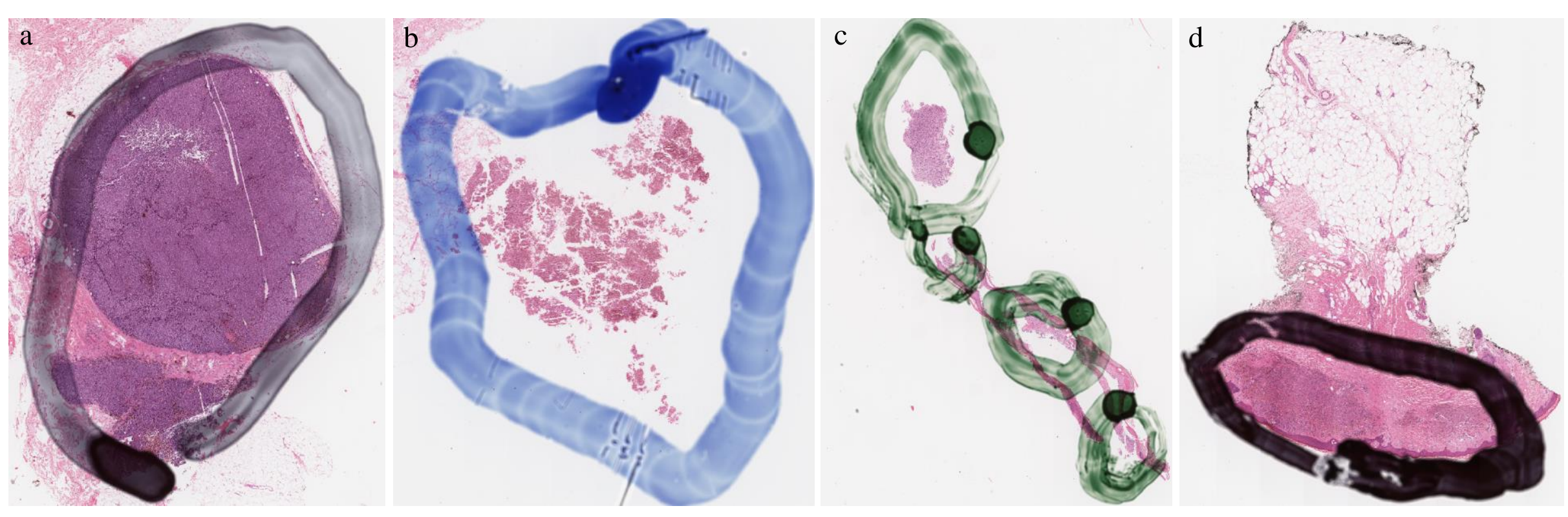}}
\caption{
Dataset was divided into 4 categories based on marker ink color and transparency: a) black, b) blue, c) green, and d) opaque.
}
\label{fig:categories}
\end{figure}

We surveyed an internal dataset of 1,100 H\&E stained WSIs from human melanoma tissues. 
In 305 of these images, markers were used to delineate the tumor border with colors black, green, and blue (250, 50, and 5 WSIs respectively.)
In 170 of the images, the intensity of the ink made the underlying tissue visually imperceptible. 
Therefore, we divided the data into four categories (Figure~\ref{fig:categories}): black (80), blue (5), green (50), and opaque (170) .
None of the WSIs contained annotations with more than one marker ink color.

We trained two models using two different training sets. In the first one, we aimed for data balance by handpicking 12 WSIs (three from each category). 
In the second dataset we maximized data diversity by increasing the number of WSIs to 300, introducing an imbalance where 80\% of the occluded images contained black marker ink. 
In each case, 5 of the WSIs were set aside for testing, with two slides from the black and one from every other category.

The marker region in each of the WSIs was segmented using HistoQC toolkit \cite{HistoQC} and the segmentations were carefully corrected in ImageScope software (Aperio, ImageScope version 12.4).
Training and testing patches of size 128x128 pixels were then extracted at random locations.
Any patch with partial or full marker regions was considered a marker patch, while clean patches lacked any amount of marker ink. 
The patches from the empty background of the WSIs were included in the dataset to steer the model to learn the removal of marker ink rather than generating tissue, when the marker image was totally opaque. 
However, the number of background patches was maintained under 25\% of the total clean patches.
The total number of patches in either training dataset was 250,000 with half of them being marker patches.

CycleGAN is composed of two generative and one discriminative CNNs, which are trained adversarially.
One of the generative CNNs is trained to the remove marker ink by outputting a patch that is similar enough to a clean tissue to mislead the discriminate CNN, which is trained in detecting the originally clean patches from those output by the generative CNN.
The second generative CNN is trained to reconstruct the input using the output of the first CNN. 
Although, this synthetic marker patch is not used, this process ensures the preservation of the visual information individual to each input patch.

Originally, CycleGAN failed to converge due to the discriminative CNN overpowering the generative ones, thus depriving them of the gradient information they need to adapt to an improving discriminative model. 
To remedy this, we changed the discriminative optimizer from Adam to stochastic gradient descent with a learning rate of 0.0001. 
Adam was used for the generative models with a learning rate of 0.0002. 
No decay was applied to the learning rates.
The training was done from scratch for 150 epochs with a batch size of 64 patches on a Nvidia Quadro P6000 GPU.

\begin{figure*}
  \centering
  \centerline{\includegraphics[width=\textwidth]{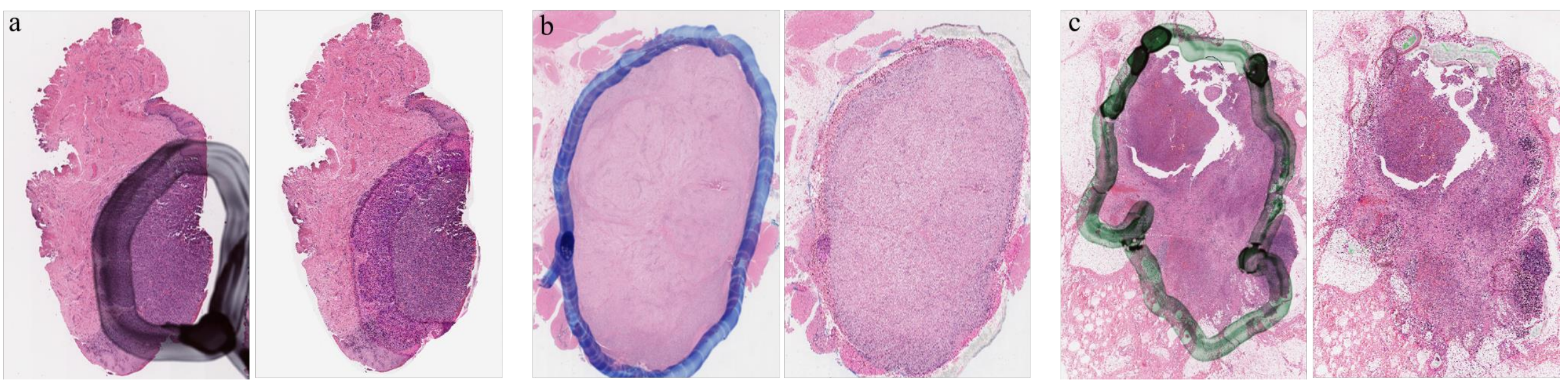}}
\caption{
Marker occluded whole slide images and reconstruction output for categories black (a), blue (b), and green (c)
}
\label{fig:results-wsi}
\end{figure*}

In the testing phase, when a reconstruction of the whole slide was needed, it was divided into 128x128 patches and each patch including marker ink was input to the generative network separately and then stitched together to reconstruct the original WSI (Figure~\ref{fig:results-wsi}). 
To avoid the resulting checkerboard alias, these patches were chosen at a stride of 100 pixels and the intensity of the overlapping pixels was averaged.
The total number of patches processed for the 5 slides was 2,000,000.

\subsection{Validation}

Our goal in this work was to correct as much of the marker regions as possible while: (1) maintaining the tissue structure underneath the marker; and (2) ensuring that the corrected regions look indistinguishable  from uncontaminated H\&E regions to the downstream human- or computer-guided analysis.
We designed four validation experiments to quantify the fulfillment of the aforementioned goals.

To measure the success of our method in transforming the marker regions into clean H\&E, we had the corrected results tested against uncontaminated tissue once by a classifier CNN and once manually. 
The manual blind test was performed by an H\&E image quality control specialist in our organization. 100 patches of size 500x500 pixels were randomly extracted from the test set of 5 WSIs, where half of the patches were clean tissue and the rest corrected marker tissue regions. The expert was then asked to discern which of the patches were corrected and which ones uncontaminated originally.

To test the similarity of the clean and the corrected regions, we also used a binary classifier CNN. 
A ResNet~\cite{resnet} with 50 layers was trained on 124,000 clean and marker image patches with a validation and test size of both 41,000 patches. 
The patches were extracted from the same 12 WSIs used in training the CycleGAN as the aforementioned strategy. 
The training was performed for 100 epochs with a batch size of 128 patches with a learning rate of 0.0001 and an Adam optimizer.
To test the similarity of the corrected patches to the uncontaminated ones, 100 corrected non-background patches were chosen from the reconstructed test set of 5 WSIs and fed to the CNN. 
Ideally, the corrected patches look identical to the clean ones and are classified as clean tissue by the CNN. 
Therefore, the percentage of marker patches classified as clean was considered as a success measure of our method.

Historically, CycleGAN excels at translating image texture while keeping the structure intact. 
However, we needed to confirm that as the marker trace in the images is removed, the boundary of the nuclei is not altered.
To test the fidelity of the reconstructed images, we randomly sampled 120 of the test patches, 30 from each category. 
An alteration of the nuclei and other tissue borders would manifest itself through discrepancy of edge information in the two images.
To measure edge similarity, we calculated the correlation of image gradient magnitudes for each input and output image patch.

The purpose of correcting H\&E images for marker is to recover the information covered by marker ink. To assess the efficacy of our method, a nuclei segmentation algorithm was designed in Definiens Tissue Studio 4.4.2 software (Definiens AG, Munich) and the number of segmented nuclei was counted in the test set of WSIs before and after correction.

\section{Results}

\begin{figure}[htb]
  \centering
  \centerline{\includegraphics[width=.5\textwidth]{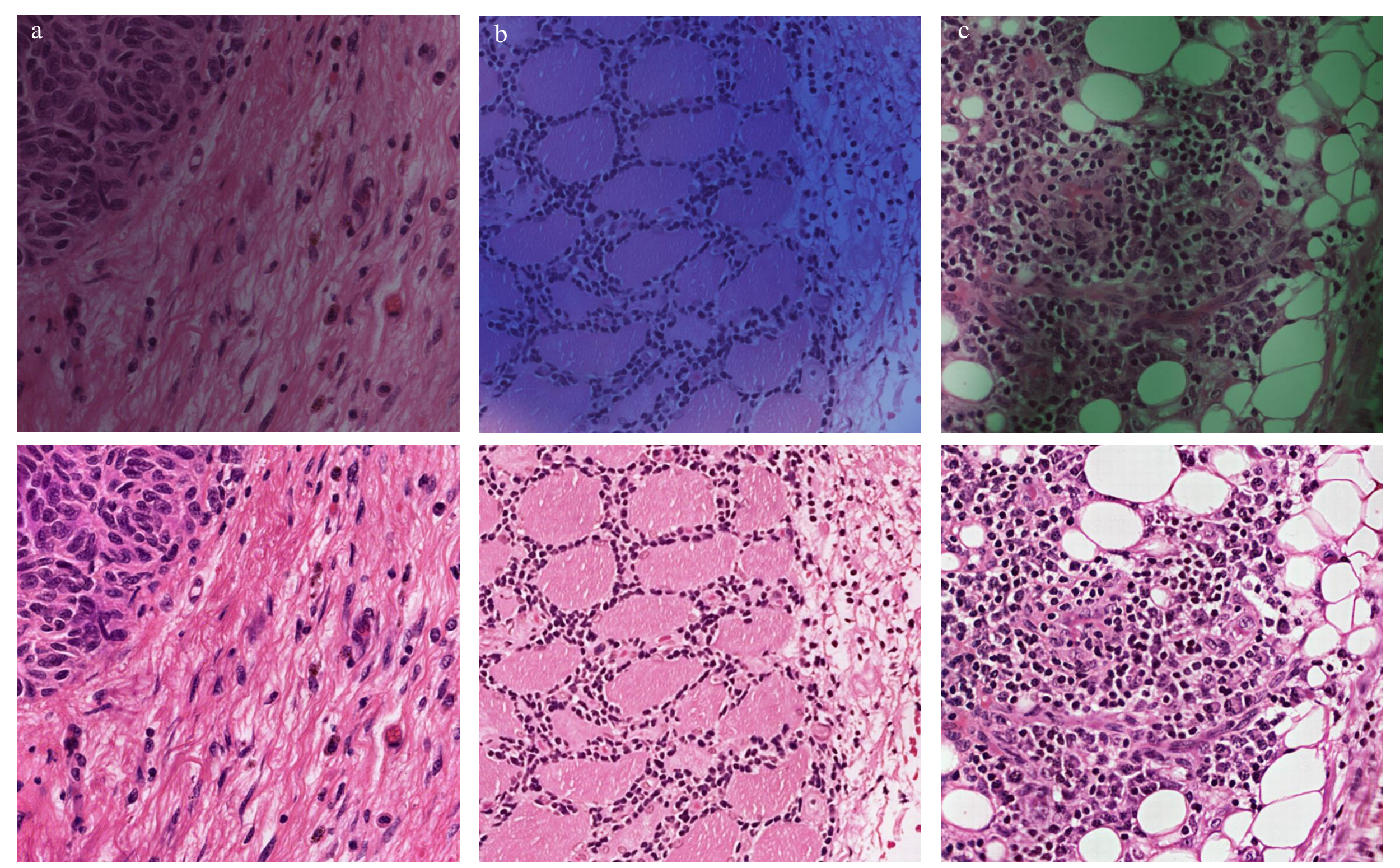}}
\caption{
Marker occluded patches (top row) and reconstruction results (bottom row) for black (a), blue (b), and green (c)
}
\label{fig:results-patches}
\end{figure}

For the model trained on the smaller dataset, more than 96\% of the marker test patches, which were corrected for marker ink, were classified as clean tissue by ResNet. 
This number was 97\% for the model trained on the larger dataset.
The breakdown was 98\% for black, 94\% for green, 96\% for blue, and 97\% for opaque categories for the first model; and 98\%, 93\%, 98\%, and 98\% for the second model.
In the blind test, the human specialist misclassified 70\% (35) of the corrected image patches as uncontaminated. 
This is while 40\% (20) of the clean images were classified as marker-corrected patches. 
We conducted the rest of the validation using the model trained on the smaller dataset. A sample of reconstruction results can be seen in Figure~\ref{fig:results-patches}.

The correlation between image gradient magnitude was an average ($\pm$STD) of 0.93 ($\pm$0.02) for non-opaque categories: 0.95 ($\pm$0.02), 0.93 ($\pm$0.02), and 0.92 ($\pm$0.03) for black, blue, and green.
In a visual examination of the image pairs, it was observed that in the images with lower correlation, either the image included the edges of the marker, or different shades of the marker constituted  marker-based gradient information.
Therefore, the removal of marker trace resulted in higher discrepancy of gradient information.
The lowest and highest correlation values were 0.97 and 0.83.
The correlation was 0.61 ($\pm$0.21) for opaque patches.
This is due to the fact that when the algorithm faces the empty content of such images, it adds random texture to the image, which make the edge correlation considerably lower than when only the ink is removed.

The number of segmented nuclei before and after correction can be seen in Table~\ref{tab:nuclei-count}.
In the five WSIs that were tested, marker ink removal resulted in the revival of 5,400 to 94,668 nuclei per image. Prior to correction, nuclei in these regions were discarded, preventing in-depth and complete analysis. 

\begin{table}
\caption{
Number of detected nuclei in whole slide images before and after correction.
}
\label{tab:nuclei-count}
\begin{center}
\begin{tabular}{cccc}
\hline
WSI & before correction & after correction & revived nuclei \\

1 & 385,314 & 461,880 & 76,566 \\
2 & 205,608 & 290,564 & 84,956 \\
3 & 130,292 & 184,489 & 5,4197 \\
4 & 314,552 & 387,201 & 72,649 \\
5 & 215,444 & 310,112 & 94,668 \\
\hline
\end{tabular}
\end{center}
\end{table}

\section{Discussion}

In this paper, we proposed a method for removing marker ink from digital H\&E images.
Markers are used by pathologists to annotate microscopic glass slides in real world. 
However, marker ink often covers important parts of the subject tissue, such as tumor margin, that contains important information for several applications.
State-of-the-art color normalization methods fail to remove marker ink as they are designed to deal with subtle variations in normal H\&E images.

We approach this problem with a state-of-the-art style transfer algorithm, CycleGAN, which is known to change texture and color without distorting image morphology. 
We showed that 70\% of the marker patches corrected by this method were indistinguishable to an originally clean image to a human expert.
Also, more than 97\% of such patches were classified as originally clean by a ResNet trained for testing purposes. 
We tested the morphology of the image after the correction and demonstrated the conservation of edge information in the process by calculating the correlation of image gradient magnitudes before and after correction.

We trained two models using two different datasets with 12, and 300 WSIs, but the same number of patches. 
The results from the latter model were only incrementally better, which has to be due to the large number of opaque markers in that dataset and the imbalance of the data compared to the former.

In our experiments, restoring the regions underneath the marker ink resulted to the revival of up to 94,000 nuclei in one WSI (20\% of all the nuclei in the slide).
In addition to the number of these revived nuclei, their biologic function makes them essential to consider in histopathology studies. 
Since markers are often used to delineate the tumor on a slide, they usually cover nuclei on the margins of the tumor, which are especially relevant in studies such as the assessment of immune response for immunotherapy.
When facing uniform regions, such as background, marker inked background, or absolutely opaque marker regions, the model generates random texture, which is easy to detect due to the low edge correlation of the input and output. 
Therefore, detection of such regions could be performed simultaneously to the correction process.

Our method is currently used in our labs by different pathology teams. Future improvements to the model will be made based on the feedback from pathologists using our results.

%


\bibliographystyle{IEEEbib}
\bibliography{strings,root}

\begin{thebibliography}{10}

\bibitem{io1}
Shousheng Liu, Pengfei Kong, Xiaopai Wang, Lin Yang, Chang Jiang, Wenzhuo He,
  Qi~Quan, Jinsheng Huang, Qiankun Xie, Xiaojun Xia, et~al.,
\newblock ``Tumor microenvironment classification based on t-cell infiltration
  and pd-l1 in patients with mismatch repair-proficient and-deficient
  colorectal cancer,''
\newblock {\em Oncology letters}, vol. 17, no. 2, pp. 2335--2343, 2019.

\bibitem{io2}
Xianjie Jiang, Jie Wang, Xiangying Deng, Fang Xiong, Junshang Ge, Bo~Xiang,
  Xu~Wu, Jian Ma, Ming Zhou, Xiaoling Li, et~al.,
\newblock ``Role of the tumor microenvironment in pd-l1/pd-1-mediated tumor
  immune escape,''
\newblock {\em Molecular cancer}, vol. 18, no. 1, pp. 10, 2019.

\bibitem{io3}
Christian~U Blank, John~B Haanen, Antoni Ribas, and Ton~N Schumacher,
\newblock ``The “cancer immunogram”,''
\newblock {\em Science}, vol. 352, no. 6286, pp. 658--660, 2016.

\bibitem{CycleGAN}
Jun-Yan Zhu, Taesung Park, Phillip Isola, and Alexei~A Efros,
\newblock ``Unpaired image-to-image translation using cycle-consistent
  adversarial networks,''
\newblock in {\em Proceedings of the IEEE international conference on computer
  vision}, 2017, pp. 2223--2232.

\bibitem{fu2018fluorescence}
Chichen Fu, Soonam Lee, David~Joon Ho, Shuo Han, Paul Salama, Kenneth~W Dunn,
  and Edward~J Delp,
\newblock ``Fluorescence microscopy image segmentation using convolutional
  neural network with generative adversarial networks,''
\newblock {\em arXiv preprint arXiv:1801.07198}, 2018.

\bibitem{mahmood2018deep}
Faisal Mahmood, Daniel Borders, Richard Chen, Gregory~N McKay, Kevan~J
  Salimian, Alexander Baras, and Nicholas~J Durr,
\newblock ``Deep adversarial training for multi-organ nuclei segmentation in
  histopathology images,''
\newblock {\em arXiv preprint arXiv:1810.00236}, 2018.

\bibitem{staingan}
M~Tarek Shaban, Christoph Baur, Nassir Navab, and Shadi Albarqouni,
\newblock ``Staingan: Stain style transfer for digital histological images,''
\newblock in {\em 2019 IEEE 16th International Symposium on Biomedical Imaging
  (ISBI 2019)}. IEEE, 2019, pp. 953--956.

\bibitem{xu2019gan}
Zhaoyang Xu, Carlos~Fern{\'a}ndez Moro, B{\'e}la Boz{\'o}ky, and Qianni Zhang,
\newblock ``Gan-based virtual re-staining: A promising solution for whole slide
  image analysis,''
\newblock {\em arXiv preprint arXiv:1901.04059}, 2019.

\bibitem{HistoQC}
Andrew Janowczyk, Ren Zuo, Hannah Gilmore, Michael Feldman, and Anant
  Madabhushi,
\newblock ``Histoqc: An open-source quality control tool for digital pathology
  slides,''
\newblock {\em JCO clinical cancer informatics}, vol. 3, pp. 1--7, 2019.

\bibitem{resnet}
Kaiming He, Xiangyu Zhang, Shaoqing Ren, and Jian Sun,
\newblock ``Deep residual learning for image recognition,''
\newblock in {\em Proceedings of the IEEE conference on computer vision and
  pattern recognition}, 2016, pp. 770--778.

\end{thebibliography}

\end{document}